# Utilizing Static Analysis and Code Generation to Accelerate Neural Networks

**Lawrence McAfee** LCMCAFEE@STANFORD.EDU
**Kunle Olukotun** KUNLE@STANFORD.EDU
Stanford University, 450 Serra Mall, Stanford, CA 94305

## Abstract

As datasets continue to grow, neural network (NN) applications are becoming increasingly limited by both the amount of available computational power and the ease of developing high-performance applications. Researchers often must have expert systems knowledge to make their algorithms run efficiently. Although available computing power increases rapidly each year, algorithm efficiency is not able to keep pace due to the use of general purpose compilers, which are not able to fully optimize specialized application domains. Within the domain of NNs, we have the added knowledge that network architecture remains constant during training, meaning the architecture's data structure can be statically optimized by a compiler. In this paper, we present SONNC, a compiler for NNs that utilizes static analysis to generate optimized parallel code. We show that SONNC's use of static optimizations make it able to outperform hand-optimized C++ code by up to 7.8X, and MATLAB code by up to 24X. Additionally, we show that use of SONNC significantly reduces code complexity when using structurally sparse networks.

## 1. Introduction

Neural networks (NN) have gained much renewed interest in recent years, as they have been shown to outperform many application-specific machine learning algorithms across several domains (Bengio, 2009). Given their potential promise for helping to move the field of machine learning towards true artificial intelligence, recent research trends have shown researchers' eagerness to test NNs on larger datasets (Raina, 2009; Cai et al., 2011). However, due to the core linear algebra routines that compose most applications, NNs are becoming increasingly limited by the amount of available computational power. In cases where large datasets are desired, researchers typically resort to structurally sparse networks, which commonly refers to networks with either dense local receptive fields (Bengio & Lecun, 2007) or non-dense receptive fields (Coates & Ng, 2011). However, in order to make larger scale networks run efficiently, researchers often find themselves needing to have expert systems knowledge to build their applications.

To the benefit of algorithm efficiency, available computational power increases each year. This benefits many applications in general, but the relative efficiency increase many applications see is nowhere near as fast as the pace of hardware advancement. This is due to the fact that for most applications, programmers utilize general purpose compilers to perform much of the optimization work such that the programmer can continue to focus on the higher-level issues that their domain requires. However, general purpose compilers (e.g., GCC) are not capable of fully optimizing specialized application domains. Some general purpose platforms are more specialized to certain domains, such as MATLAB for linear algebra-based development, and are better suited for many routines that are used to compose NNs. However, as NN applications get more complex, even a platform such as MATLAB is no longer optimal due to a lack of domain specific knowledge about the underlying data structures.

Within the domain of NNs, one piece of domain specific knowledge that can be used to increase efficiency is knowing that most network architectures – where the *architecture* is defined by the choices for layer sizes, mini-batch size, and interlayer connectivity – do not change during training. This means that the data structures used to store the network architecture are capable of being statically optimized, and then generated code can be made to run the specific architecture as efficiently as possible.

In this paper, we present SONNC (pronounced "sonic"), a Statically Optimizing Neural Network Compiler[1]. The main contributions of this paper are:

– SONNC, a neural network compiler that focuses on statically analyzing and optimizing NNs, and generates efficient parallel C++ code.

– We demonstrate analyses and optimizations which use NN domain-specific knowledge.

– We demonstrate the conciseness of code utilizing SONNC by using SONNC's front-end interface to MATLAB.

– We show that SONNC without any explicit performance tuning, outperforms hand-optimized C++ code by 3.3X–7.8X, and MATLAB code by 9.2X–24X.

## 2. Related Work

Several machine learning (ML) development platforms have been introduced recently to help with scaling to larger applications. A few popular platforms are OptiML (Sujeeth et al., 2011), Theano (Bergstra et al., 2010), and GraphLab (Low et al., 2010). OptiML is a domain-specific language for ML built on top of the heterogeneous computing platform Delite (Chafi et al., 2011). It provides abstractions to allow a programmer to develop ML applications, while Delite implicitly takes care of parallelizing and running the application across multiple CPUs and GPUs. Theano is a compiler for symbolic mathematical expressions. Although meant to be a general symbolic compiler, Theano is designed to handle ML applications. Programmers develop their applications using Python and Numpy data types, and Theano implicity generates C++ and CUDA. GraphLab provides a parallel abstraction similar to MapReduce for running large scale machine learning applications on a cluster.

Similar to SONNC, each of these platforms provides useful abstractions for programming large scale machine learning algorithms. These platforms are designed to optimize ML applications in general by using optimized routines and data structures. Unlike SONNC, however, none of these other platforms focus on statically optimizing NN data structures. Neural networks are a quickly growing field within ML, and many NN applications are composed of very computationally expensive operations. SONNC aims to provide these additional abstractions and static optimizations to allow NNs to run more efficiently.

[1]Code available at: `http://github.com/sonnc/sonnc`

## 3. System Overview

SONNC makes it easier for an end user to continue scaling applications without considering the complexities of tuning high performance code. As an example, if a user wants to design a network that uses structurally sparse connectivity – either locally dense or unstructured sparsity – a great deal of development effort would need to go into developing a sparse data structure that is efficient for indexing and updating the nonzero values in the weight matrix. When using SONNC, however, the user only needs to define the network's connectivity at the beginning of her code, and the rest of the code remains unaffected by the underlying data structure. This way, the programmer only needs to focus on algorithmic *intent* rather than worry about the details of *implementation*.

In addition to optimizing for data structures of the NN application, SONNC also optimizes the algorithm's operations by transforming and condensing sequences of routines into more efficient routines. NNs typically have a very straighforward data flow, with minimal high-level control structures. This makes it possible to perform alterations on the execution graph to make the algorithm more efficient by improving caching and reducing overhead.

### 3.1. Compiler Stages

The following sections briefly overview each stage of the compiler, which include building an execution graph, analyzing and optimizing the data structures and operations, and then generating efficient parallel code.

#### 3.1.1. Building an Execution Graph

SONNC is a standalone compiler, rather than a new programming language. As such, supported data types and operations must be embedded into an existing language to allow a user to use the system in a natural way. (See Section 5.1 for a description of the currently available data types in MATLAB.) Once an algorithm is written, an additional compilation function must be called to let SONNC perform its optimizations.

SONNC supports two high-level matrix data types: a dense matrix type and a sparse matrix type. A dense matrix is used to denote a variable where any element can contain a nonzero value. A sparse matrix, however, denotes a variable whose sparsity structure does not change after initialization. In practice, the sparse type is typically only used for weight data in a NN. In addition to these two matrix types, vector and scalar types are supported. All of the standard linear algebra operations between matrices and vectors are supported that are commonly used in NN algorithms, including multiplication, elemental (e.g., dot) operations, norms, and non-linearity operations.

When SONNC's data types are connected together via the supported operations, an execution graph of the NN application is implicitly built. This graph contains the flow of operations necessary to compute the nodes at the output (i.e., weight and bias updates) from the nodes at the input (i.e., the training set).

#### 3.1.2. Analysis and Optimization

**Static graph optimizations**. SONNC performs several common static compiler optimizations, including dead code elimination, operation re-writing, subexpression elimination, and method fusion. An example of subexpression elimination is the pre-computing of all-constant-input operations. For example in the iterative shrinkage thresholding algorithm (ISTA), the algorithm repeatedly runs the update expression:

$$Z = h_{(\alpha/L)}(Z - \frac{1}{L}(ZW^T - X)W)$$

The only variable being updated in this expression is $Z$, the approximation to the sparse codes. Hence, to speed up the update expression, we can expand out the expression and precompute $W^TW$ and $XW$ such that time is not wasted repeatedly computing these constant values.

Method fusion is an important optimization for attaining high performance. The compiler scans the execution graph for recognized operation sequences, and replaces them with more concise and efficient operations that typically have better caching and less overhead. This is a place where having domain specific knowledge becomes very useful; there are many operation sequences that are shared between various NN algorithms. For example, restricted Boltzmann machines, autoencoders, and backpropogation networks all share an operation sequence of matrix multiplication followed by bias addition followed by a nonlinearity. SONNC would recognize this sequence and convert it into its own internal operation, which in the case of a sigmoid nonlinearity would be called *MultBiasSigm*. Since the bias and nonlinearity operations must be applied to each element of the preceding data matrix, significant savings can be made if these operations can be performed while the data is still in the CPU cache immediately following the matrix multiplication. SONNC contains several operation sequences that it can recognize and replace with more efficient routines.

**Data structure optimization**. Since the NN architecture does not change during training, we can parameterize the underlying data structures such that the generated code is optimized to run as efficiently as possible for the specific network architecture. The number of threads is also chosen during this stage of the compiler. The entire data structure optimization process will be described in greater detail in Section 4.

#### 3.1.3. Multithreading and Code Generation

Once the number of threads is chosen in the previous stage, the graph is expanded into a multithreaded graph where each node represents an operation performed by a single thread. Thread synchronization points are determined during this phase, and this is the last internal representation of the application before code generation. C++ code is then generated to perform the NN application.

## 4. Data Structure Optimization

Data structure optimization has the single biggest impact on performance in comparison to the other optimizations that SONNC performs. A network's architecture, again, is defined by choices for the layer sizes, mini-batch size, and interlayer connectivity. During this stage, the underlying data structures are parameterized to run efficiently for the specific application.

The number of threads is also chosen during this stage of optimization. Although not directly a parameter that affects the underlying data structures, the number of threads must be chosen jointly with the matrix blocking size (described below) in order to yield good parallel performance. Choosing the right number of threads can have a large impact on performance. The optimal number of threads varies significantly based on matrix dimensions and connectivity structures. For example, even with the same matrix dimensions, the optimal number of threads between a network that uses dense local receptive fields and a network that uses non-dense local receptive fields can vary by a factor of two or four.

### 4.1. Underlying Data Structure

Although SONNC contains the two high-level matrix data types described in Section 3.1.1 (i.e., a dense type and a sparse type), the system contains several *underlying* data structures including a dense structure, locally dense sparse structure, a few general sparse struc-tures, and a hybrid sparse structure. Each of these underlying structures are appropriate for different circumstances, and the compiler chooses which to use for each matrix within a target application. The choice of an underlying data structure is not always intuitive. For example, when a user defines a network with dense local receptive fields, a logical choice for the underlying data structure might be to use a locally dense sparse data structure, which stores information on the locations of rectangular dense blocks within a sparse matrix. In many cases, this is the best data structure to use for the application. But this is only true when the receptive field dimension is large enough. When the receptive field is small (e.g., less than about 5 x 5), the overhead of performing small dense matrix multiplications actually increases above the simpler general sparse structure. With small enough receptive fields, the general sparse structure can outperform the locally dense structure by 1.5X–2X.

### 4.2. Data Structure Parameterization

In addition to choosing the correct underlying data structure, each of these data structures is parameterizable, effectively making a wide range of different underlying structures to choose from. The two most important of these parameters are the matrix blocking size and the data layout in memory. These parameters apply to both dense and sparse data structures. The blocking size determines how the matrix's data is partitioned in memory by splitting up the matrix into separate square blocks. Smaller block sizes increase concurrency, but also increase overhead in reading and writing the matrix data. The data layout parameter sets whether matrix elements are stored in memory using row-major order, column-major order, or another format. Both the blocking size and data format significantly impact cache reuse. While the block size is typically set globally for all matrices, the data format is set individually for each variable and depends heavily on the operations and neighboring variables (in the execution graph) that directly interact with a variable. One important point to note is that the compiler must have knowledge of the L1 and L2 cache sizes in order to properly set the blocking size. In SONNC's current implementation, it implicitly discovers these values during installation, which is described in the next section.

### 4.3. Joint Parameter Selection

SONNC's ability to properly choose the underlying data structure, matrix parameterization, and number of threads represent the most important aspect of SONNC as a statically optimizing NN compiler. Properly tuning these parameters can give up to two orders of magnitude difference in performance. The joint impact of these parameters is non-linear, and so the heuristics used to optimize this stage are critical to getting good performance. To perform this tuning process, SONNC initially must run several timing tests during its installation in order to calibrate to the CPU. SONNC times matrix multiplications for several matrix dimensions, block sizes, and number of threads in order to create a large lookup table. To keep this lookup table from being too large, parameter values are swept over exponentially, and matrix dimensions are only tested up to 10,000, block sizes up to 1,000, and number of threads up to 32. For applications with matrix dimensions larger than this, timing becomes more easily predictable from the lookup table. The timing values in this table are generally nonlinear due to caching. They are also nonconvex as a function of the number of threads. Currently, SONNC uses linear interpolation between data points in order to choose parameter values for a specific application.

This tuning process is also an ongoing area of active research for the compiler. Future plans for the tuning process include training a deep learning algorithm on the parameter space in order to better learn the nonlinearities. For the current implementation, however, linear interpolation has shown to work very well when using power-of-2 spacing when creating the lookup table.

One other important point to note is that this parameter selection operation is very fast. Using the parameter values mentioned previously during the installation phase, SONNC builds a lookup table that is stored as a 160MB file which is loaded into memory during each use. When a new application is being optimized, SONNC simply interpolates the neighboring matrix settings from the lookup table to set the block size and thread count. Since this operation only includes linear array scanning and vector averaging, it required around 2-2.5 seconds to perform parameter selection. While more sophisticated and computationally expensive methods were tested, linear interpolation worked well in practice.

## 5. Productivity

One of SONNC's goals is to make it easier to write concise and expressive code, while attaining the performance of optimized C++ code. This way, programmers can focus primarily on the algorithmic intent of their applications. However, this is often not possible with nontrivial data structures, such as when network architectures contain sparse interlayer connectivities.

For example, if a LRF network is being defined, a programmer has a choice to use either a custom data structure or MATLAB's *sparse* structure. Unfortunately, either option would require additional hand-coded routines for efficient random indexing. This in turn increases code complexity significantly.

When using SONNC, however, the only difference between whether a user would like to use dense, locally dense, or unstructured sparse connectivity is a matter of how the matrix is initialized. The remainder of the network's algorithmic description would be data structure independent. The following section gives an example of how the SONNC compiler could be used in practice.

### 5.1. MATLAB-Embedded Data Types and Operations

Although SONNC's main contribution is its powerful static optimization routines, a front-end interface for MATLAB is provided to allow end users to easily integrate the SONNC back-end into existing applications. This should in many cases automatically lead to more concise code and much higher performance for NN applications. SONNC embeds four data types into MATLAB: a Vector type, a Scalar type, a DenseMatrix type, and a SparseMatrix type.

SONNC also overloads many common symbolic operators and other methods in MATLAB such that code can be written using standard MATLAB syntax. Once a user has declared her variables using SONNC data types, much of the remainder of her code should be identical to as it would be otherwise in MATLAB. The main difference is that the body of the NN convergence loop is separated from declaration of the convergence loop construct. The body of the loop is written first, followed by a declaration of the convergence loop with its stopping criterion.

### 5.2. Example Code

Algorithm 1 shows an example use of SONNC data types inside a MATLAB script that implements a single layer LRF backpropogation network. This example highlights the use of the four embedded data types (e.g., *DenseMatrix SparseMatrix*, *Vector*, and *Scalar*), one control structure (*untilConverged*), and one other method, *runNN*, used to compile and run the application. This example demonstrates the SparseMatrix constructor being initialized with a dense Matlab matrix structure (the LRFs are stored inside a mostly zero 'dense' matrix). SparseMatrix can additionally be initialized with either MATLAB's *sparse* data structure, or a cell array that contains information of the locations and data of submatrices within a larger sparse matrix, which is useful for locally dense sparse variables.

*untilConverged* specifies the convergence stopping criterion. The convergence loop iterates until the normalized difference between successive values of the Scalar type *cost* falls below the specified tolerance. The output of *untilConverged* is a data type that simply combines the information of the looping structure and the execution graph, and is used as the input to the compiler. The code is then compiled and executed using the *runNN* method.

As can be observed in this example code, the NN's routines are not dependent on the weight matrix data structure. SONNC makes it simple to initialize data structures as desired, without needing to worry about tuning code for performance.

## 6. Performance Evaluation

This section presents performance results for a set of NN applications written in MATLAB using SONNC data types. We compare these results to hand-optimized reference implementations written using both MATLAB and C++ code. In addition, we analyze the performance improvements achievable due to SONNC's static optimizations that were overviewed in Section 3.1.

### 6.1. Methodology

We compare the performance results for three different NN applications: the restricted Boltzmann machine (RBM), the autoencoder (AE), and the iterative shrinkage thresholding algorithm (ISTA). For each of these algorithms, we use two different sparsity patterns: local receptive fields (LRF) and unstructured sparsity. These experiments were run on a machine containing two quad-core Intel Xeon X5550 2.67GHz processors and 24GB of RAM. The version of MATLAB used is R2011b 7.13. The SONNC applications are algorithmically identical to the hand-optimized MATLAB and C++ implementations. For the hand-optimized versions, we made a reasonable effort to write efficient sparse routines. To implement LRFs by hand in both MATLAB and C++, we use an array-based structure where each entry contains a dense submatrix and the index of its upper left corner. For unstructured sparsity, we use MATLAB's builtin *sparse* data structure for comparison. In C++ we use the compressed sparse block format (Buluc et al., 2009), which has several parallelization benefits. To parallelize the C++ code, we divide the work up evenly

**Algorithm 1** LRF Backprop Net (SONNC-based)

```
% MATLAB data type initialization
% Sparsity set inside a mostly-zero ...
    dense matrix
V_mat = getTrainingData();
T_mat = getTrainingTargets();
W_mat = setLocalRFs();
bias_mat = zeros(1, hidden_dim);
lr_mat = 0.01;

% MiNNCS data type initialization
V = DenseMatrix(V_mat);
T = DenseMatrix(T_mat);
W = SparseMatrix(W_mat);
bias = Vector(bias_mat);
lr = Scalar(lr_mat);

% Define a single backprop iteration
H = V * W;
H = bsxfun(@plus, H, bias);
H = sigmoid(H);
dH = (H - T) .* H .* (1 - H);

W_update = W - lr * (V' * dH);
bias_update = bias - lr * sum(dH, 1);
err = (T - H) .^ 2;
cost = sum(sum(err));

% Build array for updated variables
updates = containers.Map();
updates(W) = W_update;
updates(bias) = bias_update;

% Build execution graph
outputs = {W, bias, cost};
graph = buildGraph(outputs, updates);

% Declare convergence parameters
tol = 1e-6;
mainLoop = untilConverged(cost, tol, ...
    graph);

% Execute backprop net
runNN(mainLoop);
```

over the number of threads in the processor.

In the following discussion, the SONNC-based code, hand-optimized MATLAB, and hand-optimized C++ code will be simply referred to as SONNC, MATLAB, and C++ code, respectively.

One important point to note for using MATLAB's *sparse* data structure is that writing to only the nonzero elements as the result of a matrix multiplication is an inefficient process. This significantly impacts performance for AEs and RBMs for the unstructured sparsity experiments. While these results are included for completeness, a more fair comparison is to the C++ implementation in this case.

Timing was only performed between the lines of code immediately before and after the convergence loop, so as not to be affected by initialization procedures. Each application was run 10 times using 100 iterations of the convergence loop in order to smooth out any fluctuations due to caching and other variables. We present here the averaged time of the last five executions.

### 6.2. Performance Comparison

Figures 1–4 show the performance comparison between the SONNC, MATLAB, and C++ implementations. The reported speedup is relative to the hand-optimized version in each case. In each experiment, the SONNC code runs significantly faster than either of the hand-coded implementations.

SONNC shows the most benefit for the AE and RBM with unstructured sparsity, attaining over 200X speedup over MATLAB in some cases (Figure 1). This is because, as mentioned above, updating the nonzeros of the MATLAB *sparse* structure is an inefficient process. In the other tests, SONNC yields around 9X–24X speedup over MATLAB. In contrast to the AE and RBM, ISTA obtains relatively modest speedup (about 10X) over MATLAB when using unstructured sparsity because in ISTA the weight matrix never needs to be updated. SONNC also performs better than the C++ code, generally yielding around 4.2X–7.8X speedup (Figure 2). The important thing to note here is that optimizations performed by SONNC – i.e., using knowledge of the sparsity structure – allow it to outperform C++ code that is optimized primarily for load balance.

When using LRFs (Figures 3–4), however, the MATLAB implementation is able to run much more efficiently for the weight updates than when using unstructured sparsity. When using locally dense sparsity, MATLAB is still able to utilize its underlying BLAS imiplementation to perform the matrix multiplications. Even when using LRF sparsity, however, SONNC is still able to yield 15X-24X speedup over MATLAB, and 3.3X–6.1X speedup over C++.

Additionally, as detailed in Section 5, SONNC is able to yield these levels of performance with much more succint code. If the user ever wants to switch their network between a dense, LRF, or unstructured sparse interlayer connectivity, it is just a matter of changing the matrix's initialization, and all the performance benefits will automatically be available due to the implicit compiler optimizations.

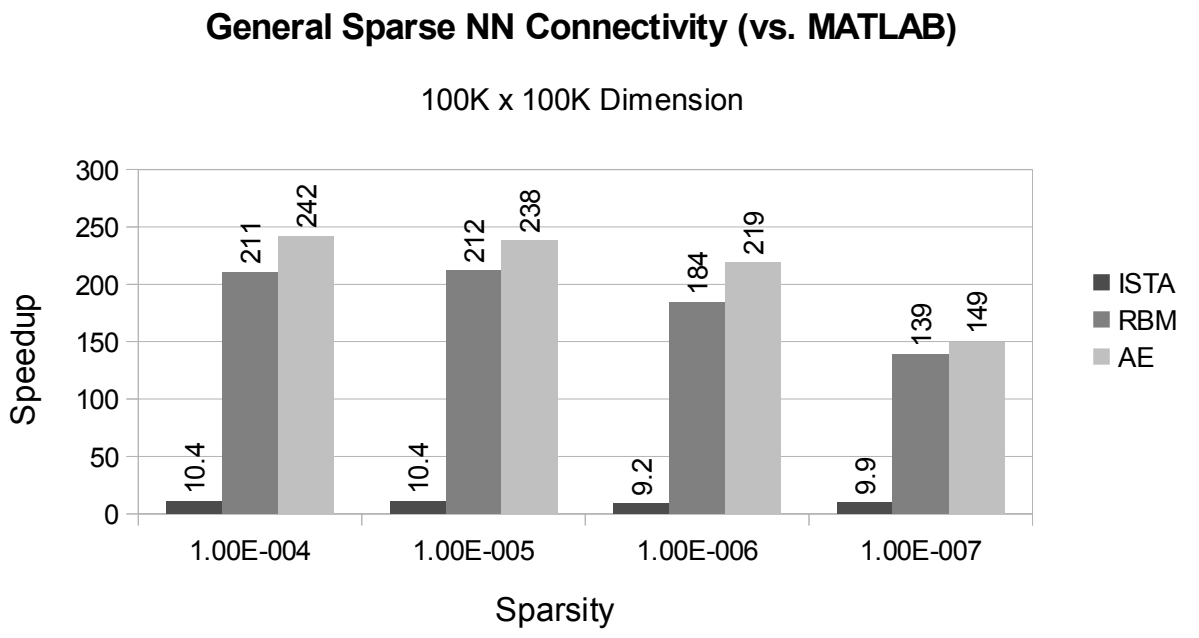

*Figure 1.* Speedup relative to hand-coded MATLAB using general sparse connectivity.

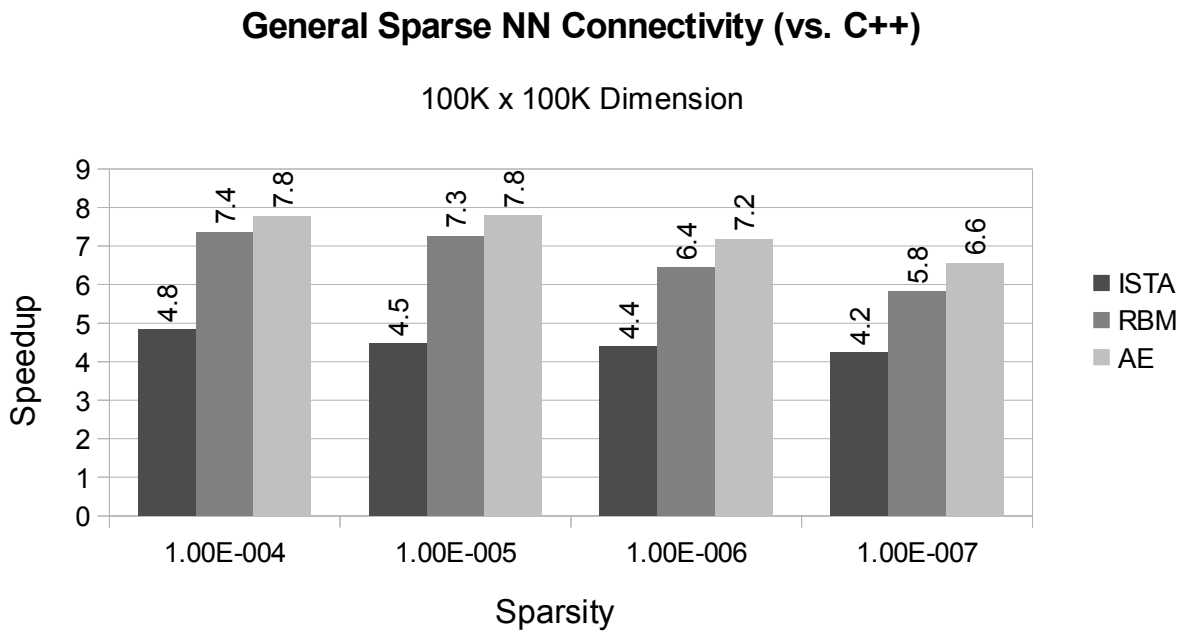

*Figure 2.* Speedup relative to hand-coded C++ using general sparse connectivity.

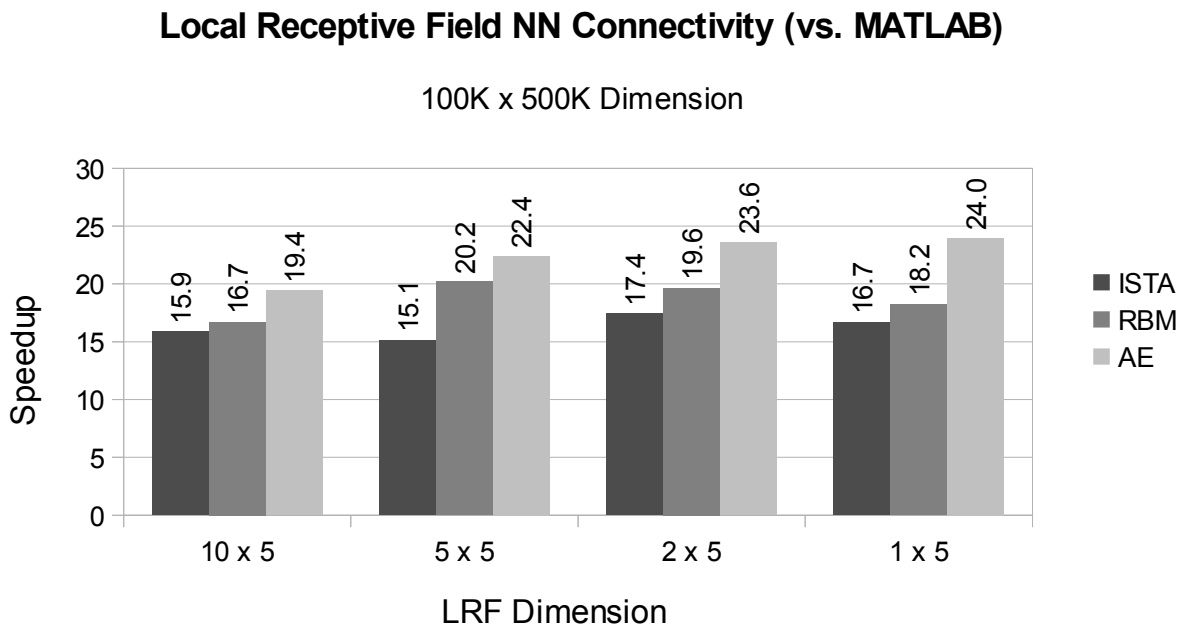

*Figure 3.* Speedup relative to hand-coded MATLAB using local receptive field connectivity.

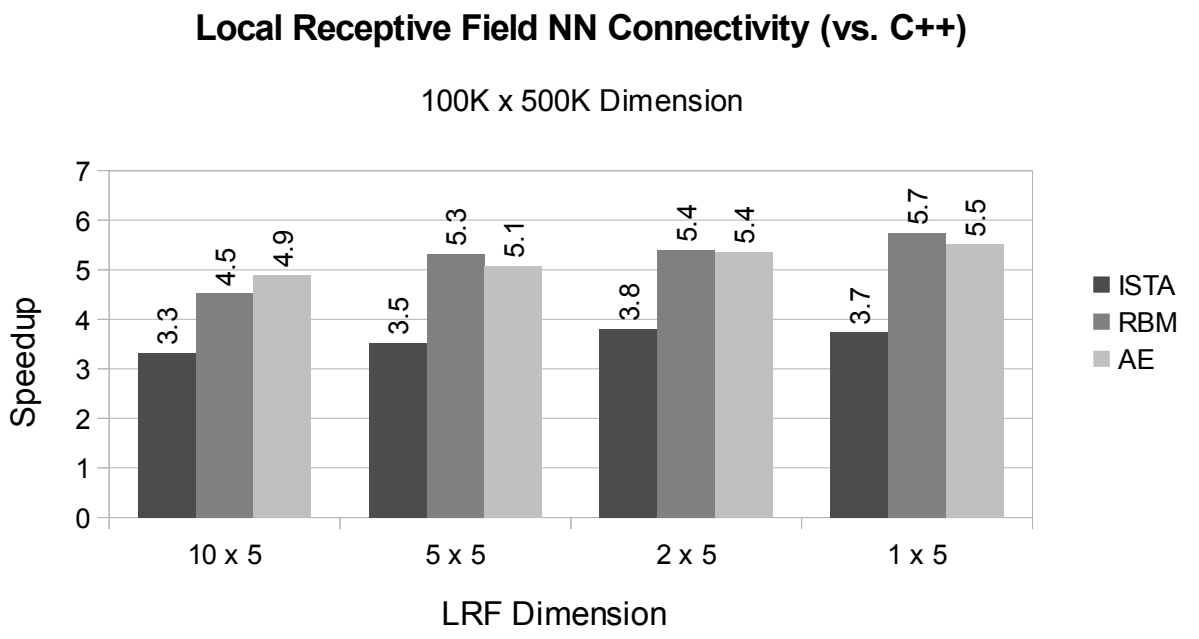

*Figure 4.* Speedup relative to hand-coded C++ using local receptive field connectivity.

### 6.3. Impact of Optimizations

Figures 5 and 6 present the impact of two of the more important optimizations described in Section 3.1. Figure 5 shows the impact of proper matrix parameterization, which includes choosing the best underlying data structure format and number of threads. Tuning these parameters has the most impact of any optimization stage. This figure demonstrates the nonlinearity of jointly tuning the matrix blocking size and number of threads. For any given block size, there is typically a single optimal setting for the number of threads. However, each block size has a different optimal setting for the number of threads, since smaller block sizes can utilize more threads. But smaller block sizes also have increasingly more overhead. This figure shows that, in this case, selecting the correct combination of these parameters has a 1.5X impact on performance for the two different block sizes.

Figure 6 shows the impact of method fusion. SONNC has a large number of common and NN-specific operation sequences that it can recognize and replace with more efficient routines. The new routines typically optimize cache reuse by combining adjacent operations into the same loop. An example of this is the *MultBiasSigm* method described in Section 3.1.2. This figure shows that method fusion is able to attain a 1.9X to 2.6X speedup for the tested algorithms. Less speedup is attained for ISTA, which is algorithmically simpler than the AE or RBM, and so has fewer operation sequences that can be fused.

## 7. Conclusion

Many promising neural network learning algorithms are facing computational challenges as they scale to larger datasets. Although the available computational power increases each year, the pace of neural network algorithmic efficiency does not advance as quickly due to the use of general purpose compilers that NN programmers rely on to optimize their applications. As NNs are applied to larger datasets, and algorithm complexity increases, application efficiency is becoming critical in order to continue advancing the field of research. In this paper, we presented SONNC, a compiler that performs static optimization of a NN application in order to generate high performance parallel code. In addition to standard compiler optimizations, SONNC relies on the domain specific knowledge that NN architecture does not change during training, which allows the compiler to optimize the underlying data structures used to store the network's architecture. We showed that SONNC was able to outperform MATLAB implementations by 9X–24X, and C++ implementations by 3.3X–7.8X. Additionally, we demonstrated how programmer productivity can be increased when using SONNC. SONNC abstracts the underlying data structure, which reduces code complexity, but still allows algorithms to attain performance better than optimized C++ code.

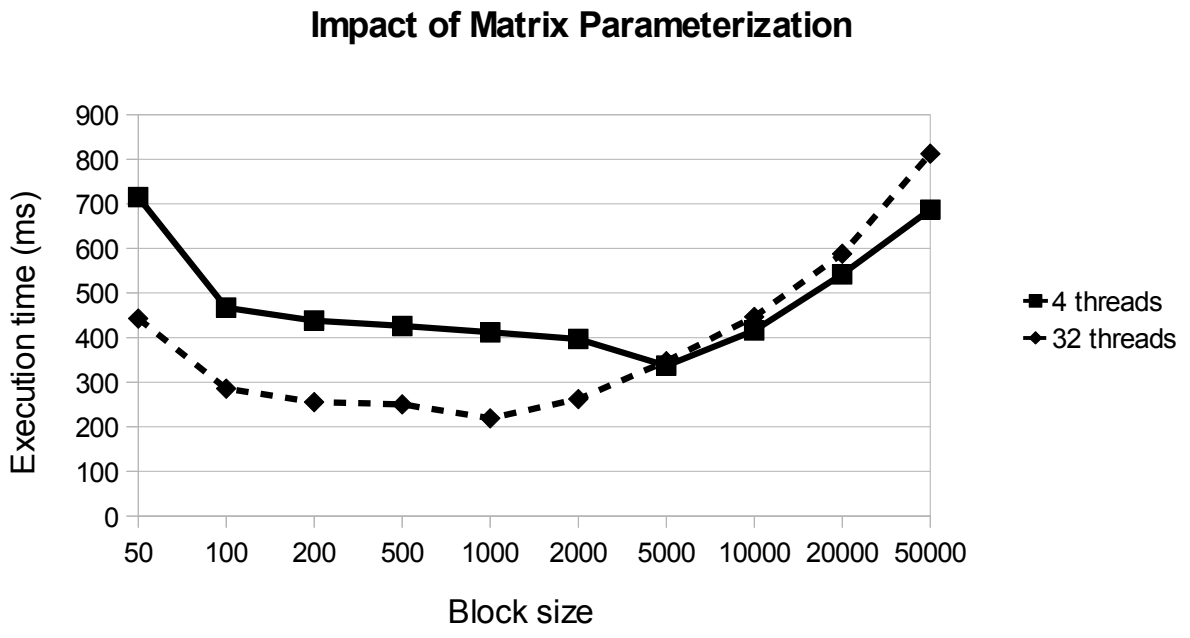

*Figure 5.* Matrix parameterization is a difficult and non-linear process. The matrix blocking size and number of threads must be chosen jointly in order to maximize performance.

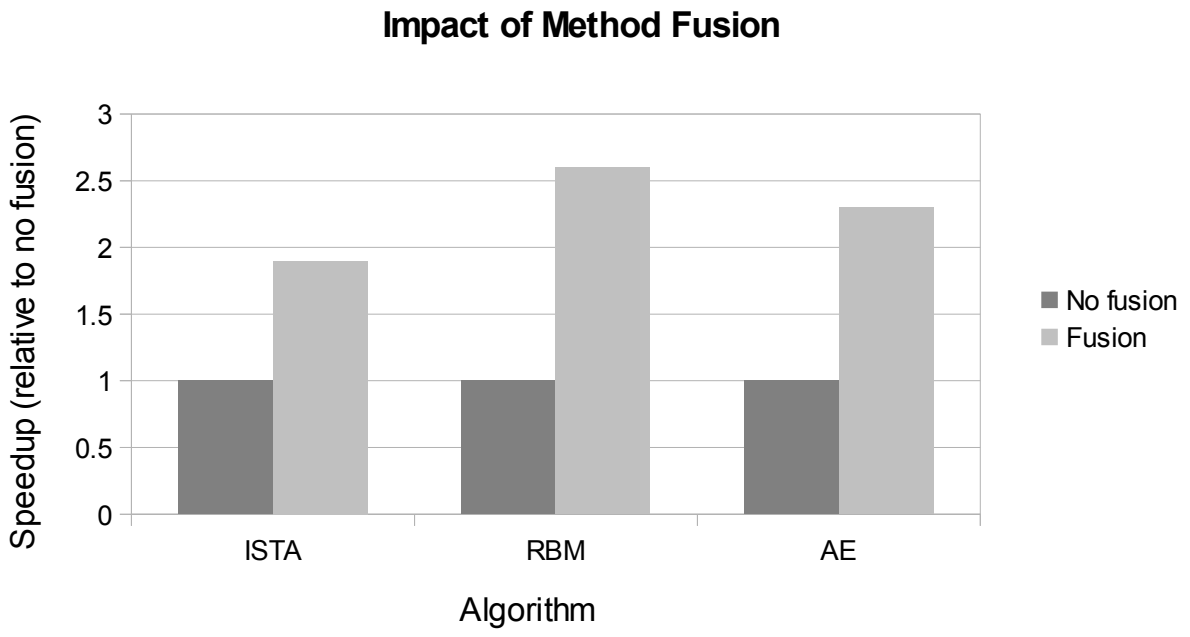

*Figure 6.* Method fusion gives a 1.9X to 2.6X performance increase. ISTA, being a simpler algorithm, does not have as many fusable operations and therefore does not benefit as much from fusion as the AE and RBM.